\documentclass{article}

\PassOptionsToPackage{numbers}{natbib}
\usepackage[preprint]{neurips_2021_arxiv}
\usepackage{array}
    \newcolumntype{P}[1]{>{\centering\arraybackslash}p{#1}}
    \newcolumntype{M}[1]{>{\centering\arraybackslash}m{#1}}

\usepackage[utf8]{inputenc} 
\usepackage[T1]{fontenc}    
\usepackage{hyperref}       
\usepackage{url}            
\usepackage{booktabs}       
\usepackage{amsfonts}       
\usepackage{nicefrac}       
\usepackage{microtype}      
\usepackage{xcolor}         
\usepackage{amsmath,amssymb,amsthm}
\usepackage{url}
\usepackage{tikz}
\usepackage{subfigure}
\usepackage{multirow}
\usetikzlibrary{spy}
\usetikzlibrary{arrows,automata}
\usepackage{pgfplots}
\usepackage{mathrsfs}  
\usepackage{enumitem}
\usepackage{wrapfig}

\pgfplotsset{width=7cm,compat=1.8}

\usepackage{algorithm, algorithmic}

\newcommand{\X}{\mathcal{X}}

\newcommand{\Y}{\mathcal{Y}}

\newtheorem{thm}{Theorem}

\newtheorem{definition}{Definition}

\DeclareMathOperator*{\argmin}{arg\,min}

\title{Learning convex regularizers satisfying the variational source condition for inverse problems}

\author{%
  Subhadip Mukherjee$^1$, Carola-Bibiane Sch\"onlieb$^1$, and Martin Burger$^2$\\
  $^1$Department of Applied Mathematics and Theoretical Physics, University of Cambridge, UK\\
  $^2$Department of Mathematics, University of Erlangen-Nuremberg, Germany\\
  Emails: \texttt{\{sm2467,cbs31\}@cam.ac.uk, martin.burger@fau.de}
}

\begin{document}

\maketitle

\begin{abstract}
Variational regularization has remained one of the most successful approaches for reconstruction in imaging inverse problems for several decades. With the emergence and astonishing success of deep learning in recent years, a considerable amount of research has gone into data-driven modeling of the regularizer in the variational setting. Our work extends a recently proposed method, referred to as \textit{adversarial convex regularization} (ACR), that seeks to learn data-driven convex regularizers via adversarial training in an attempt to combine the power of data with the classical convex regularization theory. Specifically, we leverage the variational source condition (SC) during training to enforce that the ground-truth images minimize the variational loss corresponding to the learned convex regularizer. This is achieved by adding an appropriate penalty term to the ACR training objective. The resulting regularizer (abbreviated as ACR-SC) performs on par with the ACR, but unlike ACR, comes with a quantitative convergence rate estimate.         
\end{abstract}

\section{Introduction}
Linear inverse problems seek to recover an unknown parameter $x\in\X$ from its noise measurement $y^{\delta}=Ax+e\in\Y$, where $A:\X\rightarrow \Y$ is a bounded linear operator between the Hilbert spaces $\X$ and $\Y$, and e denotes measurement noise with $\Vert e\Vert_{\Y}\leq \delta$. The clean measurement corresponds to $\delta=0$ and is denoted by $y^0$. Inverse problems are typically ill-posed, in the sense that $A$ is either non-invertible or poorly conditioned, leading to noise amplification in the solution obtained via na\"ive inversion. Variational methods \cite{benning2018modern,schuster2012regularization}, wherein one seeks to trade-off data-fidelity with a regularizer, has traditionally been the most popular approach for computing a stable solution to ill-posed inverse problems, and are rooted in a rigorous function-analytic foundation.

With the advent of deep learning, considerable attention has been devoted to leveraging the availability of data for solving inverse problems \cite{data_driven_inv_prob}. A particularly notable deep learning based endeavor has sought to model the regularizer via over-parametrized deep neural networks and learn in a data-driven manner, instead of using hand-crafted functionals \cite{nett_paper,ar_nips,acr_arxiv,kobler2020total}. Such methods can inherit the theoretical guarantees (such as stability and convergence) offered by the variational framework while effectively utilizing the power of data, provided that the regularizer fulfills certain conditions (such as convexity \cite{acr_arxiv}). This work essentially builds upon the approach introduced in \cite{acr_arxiv}. By incorporating an additional penalty term during training, our approach encourages the ground-truth images to satisfy the so-called variational source condition (SC), which leads to precise convergence rate estimates.
\section{Variational source condition}
In \cite{sc_burger} convergence rate estimates for variational reconstruction with convex regularizers were derived under the source condition, which provides an additional regularity condition on the solution of ill-posed inverse problems (see \cite{benning2018modern} for a detailed discussion). Here, we briefly recall the key results to make the exposition self-contained. Consider the variational regularization approach that minimizes an energy functional, given by
\begin{equation}
    x_{\lambda}\in\underset{x\in\X}{\argmin}\,\frac{1}{2}\Vert y^{\delta}-Ax \Vert_2^2+\lambda\,\psi_{\theta}(x).
    \label{var_recon}
\end{equation}
Here, $\left\{\psi_{\theta}\right\}_{\theta\in\Theta}$ is a convex regularizer parametrized by an input-convex neural network (ICNN) \cite{amos2017input}.
\begin{definition}[Source condition]
The $\psi_{\theta}$-minimizing solution is defined as
\begin{equation}
    \tilde{x}\in\underset{x\in\X}{\argmin}\,\psi_{\theta}(x)\,\,\text{subject to}\,\,Ax=y^0.
    \label{psi_min_sol}
\end{equation}
The variational source condition is satisfied if there exists some $\tilde{w}\in\Y$ such that $A^*\tilde{w}\in\partial\,\psi_{\theta}(\tilde{x})$.
\end{definition}
One can show  that the set of $\tilde{x}$ satisfying the SC is the same as the set of solutions to the variational problem \eqref{var_recon}, see \cite[Proposition 1]{sc_burger}. This serves as the main motivation behind our approach, in that we encourage the ground-truth images during training to satisfy the SC and thereby be the solution to the variational problem corresponding to the learned regularizer. 

Let $D_{\psi_{\theta}}(x_1,x_2):=\left\{\psi_{\theta}(x_1)-\psi_{\theta}(x_2)-\langle u,x_1-x_2\rangle\big|u\in \partial \psi_{\theta}(x_2)\right\}$ be the Bregman distance corresponding to $\psi_{\theta}$. Then, the following convergence rate can be established (see \cite{sc_burger}):
\begin{thm}
Let $\|y^{\delta}-y^0\|\leq \delta$, $\tilde{x}$ be a $\psi_{\theta}$-minimizing solution as defined in \eqref{psi_min_sol}, and suppose that the SC holds. Then, for each minimizer $x_{\lambda}$ of \eqref{var_recon}, there exists $d\in D_{\psi_{\theta}}(x_{\lambda},\tilde{x})$ such that $d\leq \lambda\frac{\|\tilde{w}\|^2}{2}+\frac{\delta^2}{2\,\lambda}$ holds. Therefore, choosing $\lambda\propto \delta$ leads to an $\mathcal{O}(\delta)$ convergence rate.  
\end{thm}
Consequently, the SC enables one to derive quantitative convergence rate estimates in terms of the Bregman distance induced by the learned convex regularizer. In contrast, \cite{acr_arxiv} shows convergence, without any quantitative rate estimate, while having to incorporate an additional Tikhonov term.
\subsection{Learning ACR with the source condition (ACR-SC)}
If the forward operator $A$ is invertible, SC dictates that $\ell_{\text{sc}}(x;\theta)=\Vert (A^*)^{-1} \nabla_x \psi(x;\theta)\Vert<\infty$ must be satisfied whenever $x$ is a solution to \eqref{var_recon}. Therefore, the smaller the quantity $\ell_{\text{sc}}(x;\theta)$ is, the more suitable $x$ would be as a solution to \eqref{var_recon}. A natural way to encourage this is to penalize $L_{\text{sc}}(\theta) = \frac{1}{n}\sum_{i=1}^{n}\ell_{\text{sc}}(x_i;\theta)$, where $x_i\sim \mathbb{P}_r$ are the ground-truth images. For non-invertible $A$, one can replace the inverse with the Moore-Penrose pseudo-inverse. Denoting by $\mathbb{P}_n$ the distribution of the undesirable images in the adversarial training framework, the overall training objective becomes 
\begin{equation}
    L(\theta) = \frac{1}{n}\sum_{i=1}^{n}\psi_{\theta}(x_i) - \frac{1}{n}\sum_{i=1}^{n}\psi_{\theta}(z_i) +\lambda_{\text{gp}}L_{\text{gp}}(\theta)+\lambda_{\text{sc}}L_{\text{sc}}(\theta).
    \label{acrsc_train_loss}
\end{equation}
Here, $x_i\sim \mathbb{P}_r$ and $z_i\sim \mathbb{P}_n$ are the ground-truth and noisy/undesirable images in the training dataset, respectively. The gradient penalty is given by $L_{\text{gp}}(\theta)=\frac{1}{n}\sum_{i=1}^{n}\left(\left\|\nabla \psi_{\theta}\left(\epsilon x_i+(1-\epsilon)z_i\right)\right\|_2-1\right)^2$, where $\epsilon\sim\text{uniform}[0,1]$, which encourages the regularizer to be 1-Lipschitz (cf. \cite{ar_nips,wgan_gp}). Notably, the training loss in \eqref{acrsc_train_loss} can be computed without direct supervision (i.e., without pairs of noisy and ground-truth images), thus offering more flexibility. For solving the variational problem \eqref{var_recon} with the learned convex $\psi_{\theta}$, one can employ a simple iterative sub-gradient algorithm for minimization. Notably, an advantage of the enforced SC is that Bregman iteration techniques are available for \eqref{var_recon} and well understood (cf. \cite{benning2018modern} for a detailed discussion). We show some preliminary promising results here (see Figure \ref{mnist_denoising}) and will further pursue this approach in future research.
\section{Numerical results}
We first show a proof-of-concept of the proposed ACR-SC method considering a denoising problem on MNIST and subsequently compare it with a number of state-of-the-art supervised and unsupervised data-driven methods for the prototypical inverse problem of computed tomography (CT) reconstruction from sparse-view projection data. In both experiments, the reconstruction quality is evaluated in terms of the peak signal-to-noise ratio (PSNR) and the structural similarity index (SSIM) with respect to the target ground-truth.     

\begin{figure*}[t]
	\centering
		\includegraphics[height=0.65in]{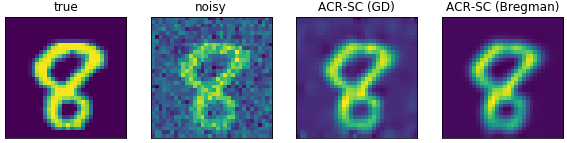}\hspace{0.5cm}
		\includegraphics[height=0.65in]{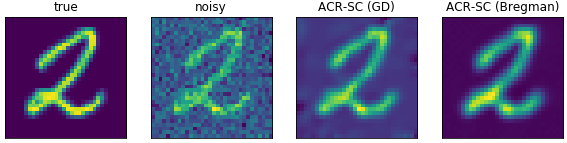}\\
		\includegraphics[height=0.65in]{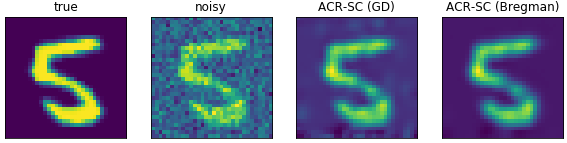}\hspace{0.5cm}
		\includegraphics[height=0.65in]{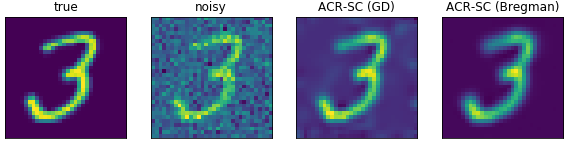}
	\caption{\small{Representative denoising examples on MNIST. The average PSNR  and SSIM over 100 randomly chosen test images are as follows: (i) noisy: $13.93\pm 0.13$ dB, $0.51\pm 0.08$ ; (ii) ACR-SC (GD): $22.72 \pm 0.64$, $0.77 \pm 0.04$; and (iv) ACR-SC (Bregman): $20.29 \pm 0.88$, $0.86 \pm 0.03$. The Bregman technique is executed with $\lambda=25$ as opposed to $\lambda=5$ in vanilla GD, and it does a comparatively better job of recovering the contrast while yielding effective denoising.}}
	\label{mnist_denoising}
\end{figure*}

For the denoising experiment on MNIST, the clean digits are corrupted with an additive white Gaussian noise with $\sigma=0.2$. The regularizer is trained over 10 epochs with a batch-size of 64 (with $\lambda_{\text{sc}}=2.0$ and $\lambda_{\text{gp}}=10.0$), to discern the clean digits from the noisy ones. Subsequently, the learned regularizer is plugged into the variational framework \eqref{var_recon}. We employ both gradient-descent (with $\lambda=5$, 300 iterations with a step-size of $0.01$) and Bregman iterations \cite{benning2018modern} (with $\lambda=25$) for solving \eqref{var_recon}. The numerical examples shown in Fig. \ref{mnist_denoising} demonstrate that ACR-SC performs a reasonable denoising and significantly improves the PSNR and SSIM over the noisy input. As shown in Fig. \ref{mnist_denoising}, Bregman iterations do a noticeably better job of circumventing loss of contrast in the reconstruction.

\begin{table}[t]
  \centering
  \begin{tabular}{l l c l r r l}
        \multicolumn{1}{l}{\textbf{method}} &
        & \multicolumn{1}{c}{\quad\textbf{PSNR (dB)}} 
        & \multicolumn{1}{c}{\quad\textbf{SSIM}} 
        & \multicolumn{1}{c}{\quad\textbf{\# param.}} 
        & \multicolumn{1}{c}{\quad\textbf{reconstruction time (ms)}} \\
        \toprule
        FBP & & $21.28 \pm 0.13$  & $0.20 \pm 0.02$ & $1$ & $37.0 \pm 4.6$\\
        TV & & $30.31 \pm 0.52$  & $0.78 \pm 0.01$ & $1$ & $28371.4 \pm 1281.5$ \\
        \midrule
        \multicolumn{5}{l}{\emph{Supervised methods}} \\
        U-Net & & $34.50 \pm 0.65$  &  $0.90 \pm 0.01$ & $7215233$ & $44.4 \pm 12.5$ \\
        LPD & & $35.69 \pm 0.60$  & $0.91 \pm 0.01$  & $1138720$ & $279.8 \pm 12.8$ \\
        \midrule
        \multicolumn{5}{l}{\emph{Unsupervised methods}} \\
        AR & & $33.84 \pm 0.63$  & $0.86 \pm 0.01$ & $19338465$ & $22567.1 \pm 309.7$\\
        ACR  & & $31.55 \pm 0.54$  & $0.85 \pm 0.01$  &  $606610$ & $109952.4 \pm 497.8$\\
        ACR-SC  &  & $31.28\pm 0.50$  &  $0.84 \pm 0.01$ & $590928$   & $105232.1 \pm 378.5$ \\
   \bottomrule
  \end{tabular}
  \\[2ex]
  \caption{\small{PSNR and SSIM statistics over 128 test slices for CT reconstruction on Mayo-clinic data. ACR-SC is competitive with ACR, uses fewer learnable parameters, and leads to convergence rate estimates.\vspace{-0.1in}}}
  \label{sparse_ct_table}
\end{table}
\begin{figure*}[t]
		\subfigure[ground-truth]{
		\includegraphics[width=1.3in]{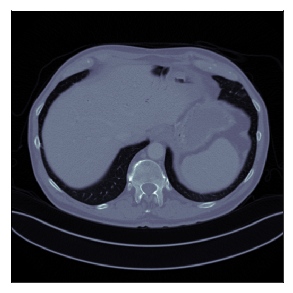}}
	\subfigure[FBP: 21.19, 0.22]{
		\includegraphics[height=1.3in]{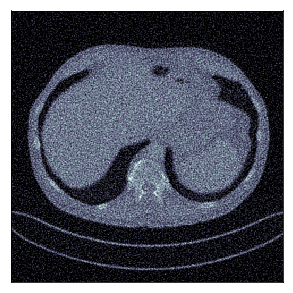}}
	\subfigure[TV: 29.85, 0.79]{
		\includegraphics[width=1.3in]{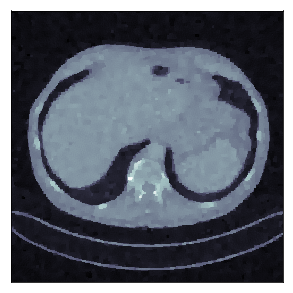}}
	\subfigure[U-net: 34.42, 0.90]{
		\includegraphics[width=1.3in]{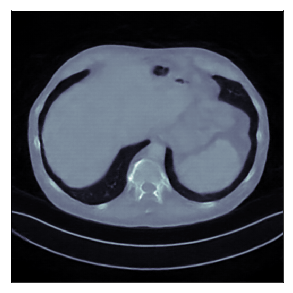}}\\
	\subfigure[LPD: 35.76, 0.92]{
\includegraphics[width=1.3in]{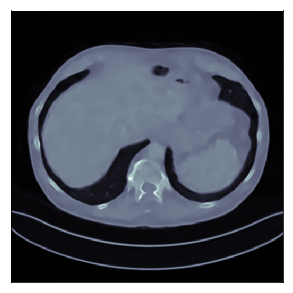}}
\subfigure[AR: 33.52, 0.86]{
		\includegraphics[width=1.3in]{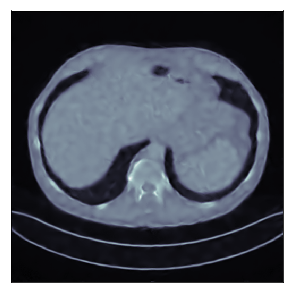}}
		\subfigure[ACR: 31.24, 0.86]{
		\includegraphics[width=1.3in]{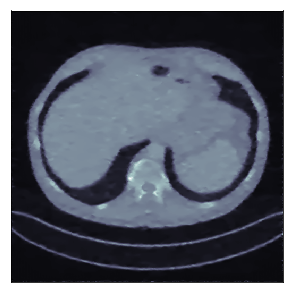}}
		\subfigure[ACR-SC: 30.93, 0.85]{
		\includegraphics[width=1.3in]{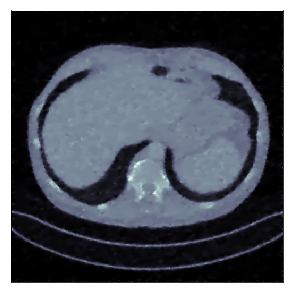}}
		\subfigure[ground-truth]{
		\includegraphics[width=1.3in]{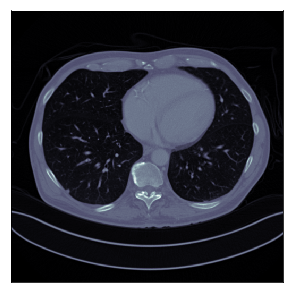}}
	\subfigure[FBP: 21.59, 0.24]{
		\includegraphics[height=1.3in]{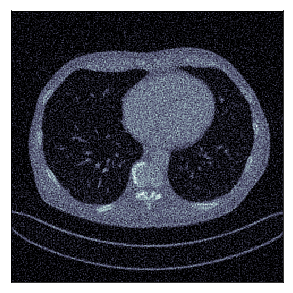}}
	\subfigure[TV: 29.16, 0.77]{
		\includegraphics[width=1.3in]{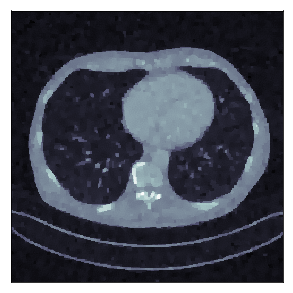}}
	\subfigure[U-net: 32.69, 0.87]{
		\includegraphics[width=1.3in]{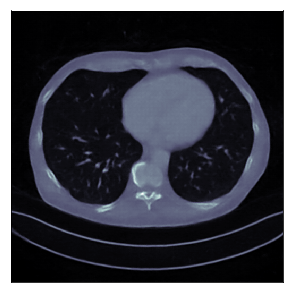}}\\
	\subfigure[LPD: 34.05, 0.89]{
\includegraphics[width=1.3in]{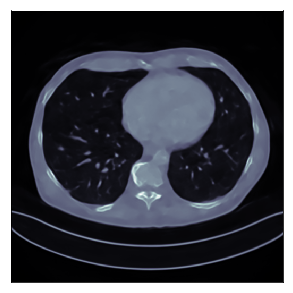}}
\subfigure[AR: 32.14, 0.84]{
		\includegraphics[width=1.3in]{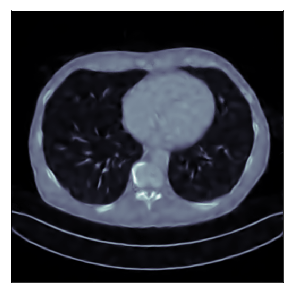}}
		\subfigure[ACR: 30.14, 0.83]{
		\includegraphics[width=1.3in]{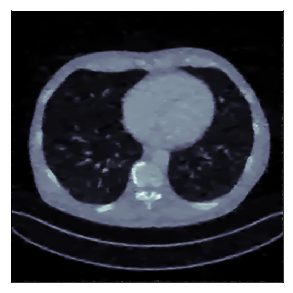}}
		\subfigure[ACR-SC: 29.88, 0.82]{
		\includegraphics[width=1.3in]{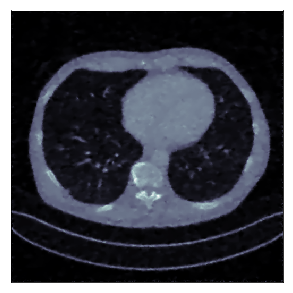}}
	\caption{\small{CT reconstruction on Mayo clinic data with the respective PSNR (dB) and SSIM scores indicated below. Enforcing the source condition leads to minor deterioration in the performance of the ACR, but we gain in terms of theoretical convergence properties.\vspace{-0.19in}}}
	\label{ct_image_figure_mayo}
\end{figure*}

For Sparse-view CT reconstruction, We adopt the same experimental setting considered in \cite{acr_arxiv} to ensure a fair comparison, and we briefly recall it here for the sake of completeness. For training the models, we use the publicly available data for the Mayo-clinic low-dose CT challenge \cite{mayo_ct_challenge}. All models are trained on 9 patients (2250 2D slices) and tested on the remaining one patient (128 slices). Parallel-beam projection data are simulated in ODL \cite{odl} with 200 uniformly spaced angles and 400 lines per angle, followed by additive Gaussian noise contamination (with $\sigma=2.0$). For comparison, we consider the classical filtered back-projection (FBP) and total variation (TV) regularization as two representative model-based approaches. Among data-driven approaches, we compare with two supervised methods, namely (i) U-Net-based post-processing of FBP \cite{postprocessing_cnn} and (ii) learned primal-dual (LPD); and two unsupervised approaches, namely (i) the adversarial regularization (AR) method introduced in \cite{ar_nips} and (ii) its convex variant ACR \cite{acr_arxiv}. Together with PSNR and SSIM, the reconstruction times are also reported for an easier assessment of the quality-vs.-time trade-off for different approaches (c.f. Table \ref{sparse_ct_table}). 

The architecture of the ICNN that models the regularizer in ACR-SC is taken to be identical to the one considered in \cite{acr_arxiv}. However, unlike \cite{acr_arxiv}, our regularizer does not have the sparsifying filter-bank and the squared-$\ell_2$ terms. We choose $\lambda_{\text{sc}}=2.0$ and $\lambda_{\text{gp}}=5.0$, and the regularizer is trained over 10 epochs with a batch-size of four. Adam optimizer with $\eta,\beta_1,\beta_2=10^{-5},0.90,0.99$ is used for training. The variational problem with the resulting regularizer is solved via gradient-descent ($\lambda=0.05$, step-size 0.8, 400 iterations). The numerical results reported in Table \ref{sparse_ct_table} and Fig. \ref{ct_image_figure_mayo} indicate that ACR-SC is only marginally inferior to ACR in terms of the quality metrics, while still outperforming classical hand-crafted regularizers such as total-variation (TV). The slight deterioration as compared to ACR can be attributed to the additional regularity constraint imposed by the SC.    
\section{Conclusions}
We addressed the problem of learning data-adaptive convex regularizers for inverse problems and proposed an approach to augment the existing adversarial learning framework with a suitable penalty that enforces the variational source condition. Incorporating the source condition penalty leads to only minor degradation in the numerical performance, but it simultaneously offers a theoretical grounding for Bregman iterations and paves the way for deriving convergence rate estimates for the resulting variational reconstruction problem as the noise-level approaches zero. 


\bibliographystyle{plainnat}
\bibliography{bib}

\end{document}